\documentclass[10pt,conference]{IEEEtran}

\usepackage[T1]{fontenc}
\usepackage[utf8]{inputenc}
\usepackage{lmodern}
\usepackage[usenames,dvipsnames]{xcolor}
\usepackage{tikz}
\usepackage{booktabs}
\usepackage{graphicx}
\usepackage[hidelinks]{hyperref}

\definecolor{acc}{RGB}{23,58,122}
\definecolor{good}{RGB}{30,107,69}
\definecolor{bad}{RGB}{154,43,34}
\definecolor{soft}{RGB}{70,76,85}

\newcommand{\axisW}{6.0}
\newcommand{\barrow}[4]{%
  \fill[#2] (0,#1) rectangle (#3*\axisW/100,#1+0.34);
  \node[anchor=west,font=\footnotesize] at (#3*\axisW/100+0.08,#1+0.17) {#4};}
\newcommand{\gridaxis}{%
  \foreach \x in {0,25,50,75,100}{%
    \draw[soft!40] (\x*\axisW/100,-0.15) -- (\x*\axisW/100,\barTop);
    \node[anchor=north,font=\scriptsize,text=soft] at (\x*\axisW/100,-0.15) {\x};}}

\begin{document}

\title{Persuaded, Not Informed:\\
Incentive-Misaligned Witnesses Defeat In-Context Grounding}
\author{\IEEEauthorblockN{Rahul Balakavi}
\IEEEauthorblockA{AmpUp Research\\ \texttt{rahulb@ampup.ai}}}
\maketitle

\begin{abstract}
Language-model agents are increasingly deployed over customer-relationship
management (CRM) records to answer operational questions such as whether a
sales lead should be qualified. We identify a failure mode not addressed by a
stronger model: when the context contains an assertion by a party with an
incentive toward optimism---here the sales representative, a witness recorded
in the CRM---the model treats the assertion as evidence and clears deals the
company's own records deem unacceptable. Across 100 lead-qualification tasks
from CRMArena-Pro, the representative asserts an acceptable timeline in every
call and an acceptable budget in 76; on the 31 tasks where such an assertion
contradicts the price list and installation policy, a model reading
only the transcript clears the deal in 29 of 31 cases. The signature is
consistent across seven models from four providers (misled on 87--97\%);
scale and explicit reasoning confer no resistance. Only 3 of 35 genuine
failures involve no assertion, so the failure is one of \emph{persuasion},
distinct from missing information. Our contribution is a diagnostic method,
not a new architecture: (i) a bucket analysis that separates persuasion from
information gaps, (ii) a same-information control demonstrating that supplying
the records to the model \emph{lowers} strict accuracy from 41 to 18 while
raising recall---precision collapses---and (iii) a compute-step control that
holds the extraction fixed and varies only who computes Budget and Timeline,
isolating the operative component. The margin ranges from 42 points on an inexpensive model to 2--5 points on models that already compute correctly; on the strongest models the arms are within confidence intervals, so the pattern is best read as a consistent direction and a soundness property rather than a proved performance floor. We pre-specify a
generalization test that returns a negative result, characterize the
precondition (policy exactly specified \emph{and} identifiable from inputs),
and release all evaluation artifacts.
\end{abstract}


\section{The Finding: Persuaded, Not Informed}
Language-model agents are increasingly pointed at company CRMs to answer
everyday questions---whether a lead is worth pursuing, whether a quote follows
policy, which deals are at risk---and benchmarks such as CRMArena-Pro
\cite{crmarena} score them against known answers. On the harder tasks these
agents underperform, and the common response is to reach for a larger model or a
more elaborate prompt. We argue that the problem lies elsewhere, and we measure
where. The failure we document sits in a small overlap between three studied
phenomena and matches none exactly. Sycophancy \cite{sycophancy, mwe, weisyc}
concerns deference to the \emph{user's} stated view; here the persuasive party
is a witness whose testimony sits in the context, not the questioner.
Knowledge conflict \cite{xie2024} concerns deference to in-context evidence
that contradicts parametric knowledge; here the model has no parametric prior
about a specific deal's price and there is no explicit conflict, only an
unverified claim. Indirect prompt injection \cite{greshake2023} and
misinformation pollution \cite{pan2023} concern adversarial content; here the
content is not adversarial, merely optimistic. We call this the
\emph{incentive-misaligned witness} case---a source recorded in the context
whose incentives are not aligned with correctness---and show that supplying the
records to the model does not neutralize it.

The task on which this is clearest is \emph{lead qualification}. Each task
provides a sales-call transcript and asks whether the lead can be qualified;
if not, which of the four factors of the BANT sales-qualification framework---\emph{Budget, Authority, Need,
Timeline}---are at fault. Two are answerable from what was said (whether the
contact has purchasing authority; whether a genuine need exists). The other two
are traps: whether the order actually fits the stated budget, and whether the
promised timeline is achievable. Both depend on the company's own catalog prices
and installation policy, not on anything contained in the transcript.

We therefore posed a sharper question than aggregate accuracy. For all 100
calls we sorted every genuine Budget-or-Timeline failure into three cases:
(a)~the representative explicitly claims the deal is acceptable while the records
say otherwise; (b)~the representative makes no claim; and (c)~the representative
flags the problem. We then measured how often a model, reading only the
transcript, misses the failure in each case. If case~(a) is large and the model
fails there, the model is being misled rather than merely deprived of data.

\begin{table*}[t]
\centering
\caption{What the representative claims versus what the records say, across 100
lead-qualification calls. The claim that the deal is acceptable is asserted
almost always; on the deals where it is false, the model accepts it.}
\label{tab:bucket}
\small
\begin{tabular}{@{}lcccc@{}}
\toprule
Dimension & Claims acceptable & Genuinely fails & \dots\ and claimed acceptable & Model misled\\
\midrule
Timeline & 100/100 & 12 & 12 & \textcolor{bad}{12/12}\\
Budget   & 76/100  & 23 & 19 & \textcolor{bad}{17/19}\\
\midrule
\textbf{Contradiction cases (pooled)} & --- & 35 & \textbf{31} &
\textbf{\textcolor{bad}{29/31 (94\%)}}\\
\bottomrule
\end{tabular}
\end{table*}

The representative asserts that the timeline is feasible in every call, and that
the budget is met in three of four. Of the 35 deals that genuinely fail on
budget or timeline, 31 are cases in which the representative claimed the
opposite; only 3 involve no claim, and 1 is a claim consistent with the records.
This excludes the mundane explanation: the model does not fail for lack of the price list---31 of 35 failed deals contain a claim in the transcript that contradicts the records. That the model then accepts those claims is a strong descriptive coincidence, not, on this design, a proved causal effect: the assertion cannot be manipulated without changing other properties of the transcript. The counterfactual test that would isolate the assertion is left as future work. A representative example: the transcript states
``that fits your budget, and we can install in a day,'' whereas the records show
the order is \$1{,}400 over budget and the installation requires three days. A
model reading only the transcript records the claim and clears the deal.

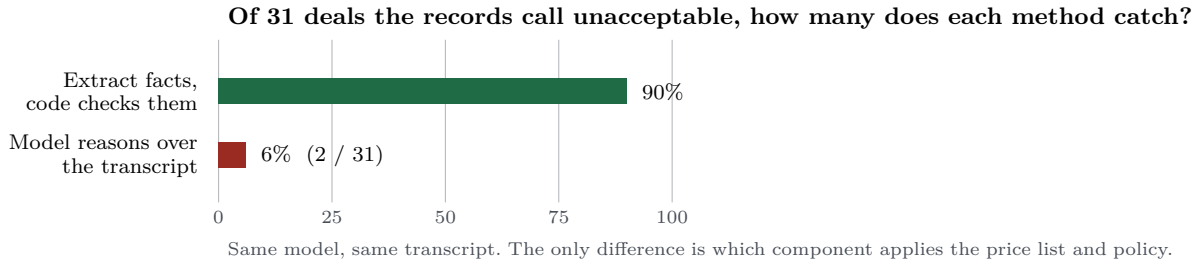
\begin{figure*}[t]
\centering
\begin{tikzpicture}
\def\barTop{2.0}
\gridaxis
\node[anchor=north west,font=\small\bfseries] at (0,2.55)
  {Of 31 deals the records call unacceptable, how many does each method catch?};
\barrow{1.15}{good}{90}{\footnotesize 90\%}
\node[anchor=east,font=\footnotesize,align=right] at (-0.15,1.32)
  {Extract facts,\\ code checks them};
\barrow{0.30}{bad}{6}{\footnotesize 6\%\ \ (2 / 31)}
\node[anchor=east,font=\footnotesize,align=right] at (-0.15,0.47)
  {Model reasons over\\ the transcript};
\node[anchor=north west,font=\scriptsize,text=soft] at (0,-0.55)
  {Same model, same transcript. The only difference is which component applies the price list and policy.};
\end{tikzpicture}
\caption{The central result. Reasoning over the transcript to an answer, a
model catches almost none of the deals the records deem unacceptable,
because the representative states that they are acceptable. Extracting the facts
and checking them in code catches nearly all of them, with no change of model or
input.}
\label{fig:contrast}
\end{figure*}

The remedy is to change the model's task rather than its size. Given the same
transcript, the model is restricted to extracting the plain facts as structured
data---the products and quantities, the stated budget, the required
timeline---after which a small amount of deterministic code checks those facts
against the catalog price and the policy. On the identical 31 contradiction
cases, this procedure catches 28; the model's own end-to-end reasoning caught 2
(Fig.~\ref{fig:contrast}).

\subsection{The Failure Is Model-Independent}
The result in Fig.~\ref{fig:contrast} is not particular to one model or one
provider. We repeated the measurement on the same 31 contradiction cases with
seven models from four providers (OpenAI, Anthropic, Moonshot, Alibaba) spanning
current-generation and prior-generation OpenAI models, Claude Sonnet~5 and Claude
Fable~5.1 \cite{claude}, Kimi~K2.6, and Qwen~3.8~Max
\cite{kimik2}. Specific model identifiers are held in the supplementary
material to keep the paper current as models turn over---reporting the fraction misled by the pitch and the fraction the
grounded procedure recovers, with 95\% Wilson confidence intervals
\cite{wilson1927,brown2001} (Table~\ref{tab:sweep}). Every model is misled on at least 87\% of the cases,
two of them (Claude Sonnet~5 and Qwen 3.8~Max) miss all but one.
Explicit reasoning confers no resistance, the frontier model is no exception,
and the effect does not depend on the provider. The grounded procedure recovers
81--90\% of the cases for every model, so both the failure and its remedy are
properties of the task rather than of a particular model. The intervals are wide because the contradiction set is small ($n=31$, the
most that this benchmark's 100 lead-qualification tasks admit); we therefore read
the table as evidence of a consistent direction rather than of precise rates. The
remainder of the paper develops why the procedure works, its cost, what component
is responsible for the result, and the boundary beyond which it does not apply.

\begin{table*}[t]
\centering
\caption{Contradiction persists across model families. On the 31 contradiction cases, the
fraction of deals each model clears despite the records (``misled by the pitch'')
and the fraction the grounded extract-then-compute procedure recovers, with 95\%
Wilson confidence intervals ($n=31$). Every model is misled on at least 87\%.
Kimi~K2.6 and Qwen~3.8~Max permit only temperature~1, so those rows are
one sample of a non-deterministic run; all others are temperature~0.
Figure~\ref{fig:computeControl} carries two older OpenAI models (GPT-4o-mini and o3-mini) that we keep for the compute-step chart because they show what code does when the model is worst at the arithmetic.}
\label{tab:sweep}
\small
\begin{tabular}{@{}llcc@{}}
\toprule
Model & Provider & Misled \% [95\% CI] & Caught \% [95\% CI]\\
\midrule
\multicolumn{4}{@{}l}{\emph{Current-generation}}\\
GPT-5 & OpenAI & \textcolor{bad}{90.3} [75--97] & 80.6 [64--91]\\
gpt-5.6-sol & OpenAI & \textcolor{bad}{90.3} [75--97] & 90.3 [75--97]\\
Claude Sonnet~5 & Anthropic & \textcolor{bad}{96.8} [84--99] & 83.9 [67--93]\\
Claude Fable~5.1 & Anthropic & \textcolor{bad}{93.5} [79--98] & 87.1 [71--95]\\
Kimi~K2.6 & Moonshot & \textcolor{bad}{87.1} [71--95] & 83.9 [67--93]\\
Qwen 3.8 Max & Alibaba & \textcolor{bad}{96.8} [84--99] & 87.1 [71--95]\\
\midrule
\multicolumn{4}{@{}l}{\emph{Prior generation (reference)}}\\
GPT-4o & OpenAI & \textcolor{bad}{93.5} [79--98] & 90.3 [75--97]\\
\bottomrule
\end{tabular}
\end{table*}

\section{The Fix: Inform the Verdict}
\begin{figure*}[t]
\centering
\begin{tikzpicture}[font=\small,
  box/.style={draw,rounded corners=2pt,minimum height=1.05cm,minimum width=2.7cm,align=center,inner sep=3pt},
  ex/.style ={draw=acc,fill=acc!8,rounded corners=2pt,minimum height=1.15cm,minimum width=2.7cm,align=center},
  cp/.style ={draw=acc,line width=1pt,rounded corners=2pt,minimum height=1.25cm,minimum width=2.7cm,align=center},
  ans/.style={draw=good,line width=1pt,rounded corners=2pt,minimum height=1.0cm,minimum width=2.3cm,align=center},
  ar/.style ={-latex,draw=acc,line width=.9pt}]
\node[box] (nl) at (0,1.5) {\textbf{NL activity records}\\\scriptsize transcript $\cdot$ email $\cdot$ case};
\node[box] (rows) at (0,-0.4) {\textbf{Structured rows}\\\scriptsize quotes $\cdot$ cases $\cdot$ orders};
\node[ex] (ext) at (3.9,1.5) {\textbf{Inexpensive extractor}\\\scriptsize text $\rightarrow$ facts (JSON)};
\node[cp] (comp) at (8.0,0.55) {\textbf{Deterministic compute}\\\scriptsize code applies the rule};
\node[ans] (a) at (12.2,0.55) {\textbf{Answer}\\\scriptsize verdict / value};
\node[box,minimum width=6.6cm] (kb) at (5.95,-2.2)
  {\textbf{Knowledge base}\\\scriptsize
   Catalog prices $\cdot$ Policies $\cdot$ Routing rules\\[1pt]
   \scriptsize\textcolor{acc}{EXACT $\cdot$ STATED $\cdot$ REUSED}};
\draw[ar] (nl) -- (ext);
\draw[ar] (ext) -- (comp);
\draw[ar] (rows) .. controls (2.6,-0.4) and (3.0,0.2) .. (comp.west|-comp.south) ;
\draw[ar] (kb) -- (comp);
\draw[ar] (comp) -- (a);
\node[font=\scriptsize\itshape,text=soft] at (2.6,0.25) {no extraction needed};
\end{tikzpicture}
\caption{The extract-then-compute pipeline. Text records pass through an
inexpensive extractor into a small structured schema; already-structured rows
bypass it. In either path the verdict is produced by code applying the
organization's exact, stated policy, rather than by a model reasoning over
that policy internally.}
\label{fig:arch}
\end{figure*}
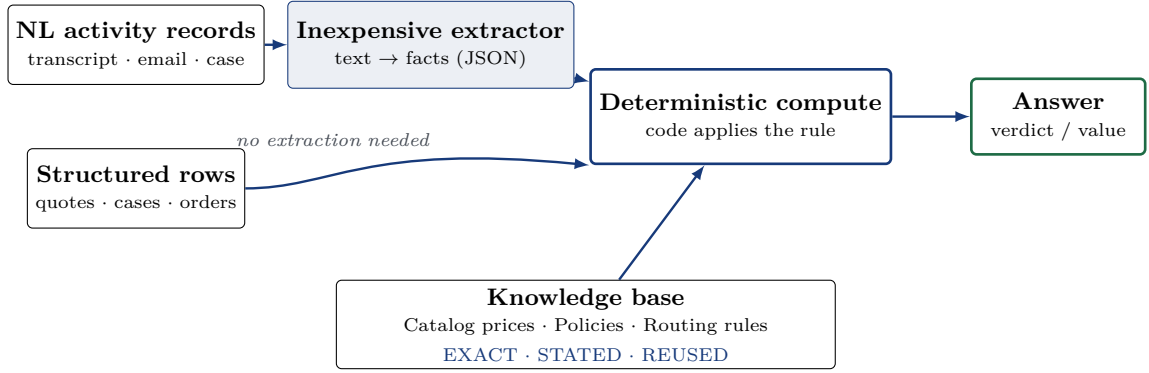

If the failure is one of persuasion, the remedy is to change the object the
model is asked to produce. We divide labor across two components under a
single rule: the model reads, code decides. The mechanism, which we call
\emph{extract-then-compute} after prior work on program-aided reasoning
\cite{pal, coc, toolformer, react}, has three parts (Fig.~\ref{fig:arch}):
(i) a knowledge base of the organization's exact, stated facts---catalog
prices, written policies---already present in the underlying database; (ii)
an extractor that converts natural-language records into a small structured
schema; and (iii) a deterministic compute step that applies the policy to the
extracted facts. When the facts are already structured rows, the extractor is
omitted. We claim no novelty for this architecture. Our contribution is
diagnostic: the controls in Section~III show that supplying the records to the
model as text does not substitute for the code path, and isolate the compute
step as the operative component.

\subsection{Cost Implications}
Answering one lead-qualification query in the extract-then-compute path is
one extraction call. The ladder (Table~\ref{tab:ladder}) shows the inexpensive
reader matches the GPT-5 within 5 points (84 versus 89), so there is
no accuracy reason to pay frontier per-query prices in this path. A tool-using
agent \cite{react} answers the same query with several frontier-priced calls
over accumulating context (schema inspection, query, rows, re-query, reason)
and, as Fig.~\ref{fig:contrast} shows, is more likely to be wrong. We do not
report absolute dollar figures; the cost ratio between the two paths is roughly two orders of magnitude on our traces (a few hundred fold at the extremes).

\section{The Controlled Comparison}
Figure~\ref{fig:contrast} contrasts two extremes. The full ladder between them
holds the input fixed and varies only the method. We evaluate on all 100 B2B
lead-qualification tasks, developing the method on the CRMArena 80/20 training
split and reporting the held-out test split. Grading is strict exact match on
the set of failing factors; where it is diagnostic we also report the looser
\emph{contain} metric (all gold factors present, extra factors tolerated), since
the two diverge in a way that exposes the failure mechanism.

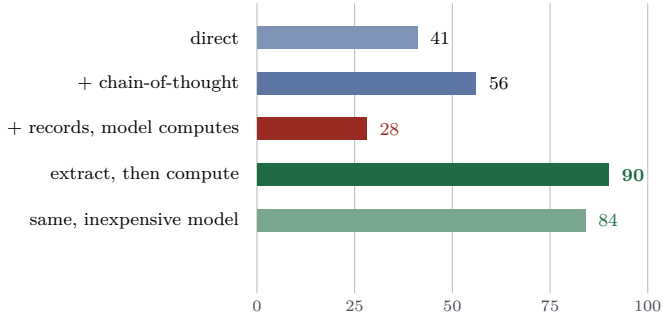
\begin{figure}[t]
\centering
\resizebox{\columnwidth}{!}{%
\begin{tikzpicture}
\def\barTop{4.3}
\gridaxis
\barrow{3.6}{acc!55}{41}{\footnotesize 41}
\node[anchor=east,font=\footnotesize] at (-0.15,3.77) {direct};
\barrow{2.9}{acc!70}{56}{\footnotesize 56}
\node[anchor=east,font=\footnotesize] at (-0.15,3.07) {+ chain-of-thought};
\barrow{2.2}{bad}{28}{\footnotesize\color{bad}28}
\node[anchor=east,font=\footnotesize] at (-0.15,2.37) {+ records, model computes};
\barrow{1.5}{good}{90}{\footnotesize\color{good}\bfseries 90}
\node[anchor=east,font=\footnotesize] at (-0.15,1.67) {extract, then compute};
\barrow{0.8}{good!60}{84}{\footnotesize\color{good}84}
\node[anchor=east,font=\footnotesize] at (-0.15,0.97) {same, inexpensive model};
\end{tikzpicture}}
\caption{Accuracy on lead qualification by method (strict set match; readers
and splits as in Table~\ref{tab:ladder}). The decisive rung is the last: applying the policy in
code. Providing the model with the records and asking it to reason makes it
less accurate under strict grading than providing nothing: the records raise
recall of true failures but the model over-flags.}
\label{fig:ladder}
\end{figure}

\begin{table}[t]
\centering
\caption{The method ladder. Train and test are the CRMArena 80/20 split;
``full'' is all 100 tasks.}
\label{tab:ladder}
\small
\begin{tabular}{@{}lcccc@{}}
\toprule
Rung --- method & Reader & Tr. & Te. & Full\\
\midrule
Direct from transcript & efficient & 44 & 40 & 41\\
+ chain-of-thought & standard & 56 & --- & ---\\
+ records, model does arithmetic & standard & \textcolor{bad}{28} & --- & ---\\
\textbf{Extract $\rightarrow$ compute in code} & standard & \textbf{88} & \textbf{90} & \textbf{89}\\
\quad same, GPT-4o-mini & efficient & --- & --- & \textbf{84}\\
\bottomrule
\end{tabular}
\end{table}

\begin{table*}[t]
\centering
\caption{Same-information control (all 100 tasks; strict shown with 95\% Wilson
intervals, $n{=}100$). Upper block: an GPT-4o-mini throughout. Strict = exact set
match; contain = all gold factors present, extras tolerated. Handing the model the records raises contain and lowers strict:
recall up, precision down. Lower block: the compute step isolated on five
models from three providers---identical extraction, only who computes
Budget/Timeline differs. Code never lowers either metric; the margin is largest
on the inexpensive model, 14--19 points pooled over two runs on the frontier
and reasoning models, shrinks to 2--4 points on Claude Sonnet 5 (three runs) and 5 on Kimi K2.6.
$^\dagger$4 Claude~Sonnet~5 responses were not parseable JSON and 5 Kimi~K2.6 calls timed out; each is graded wrong in both arms. Over parsed rows only, the pairs are
79/83 (Claude) and 79/84 (Kimi). Kimi runs at temperature~1. Confidence intervals are marginal Wilson intervals on each arm. Because both arms score the same 100 cases, the correct uncertainty for the arm-to-arm difference is a paired McNemar-style interval; the marginal intervals shown here are an upper bound on it. Per-case outcomes are released so the paired test can be computed.}
\label{tab:sameinfo}
\small
\begin{tabular}{@{}llcc@{}}
\toprule
Condition & Records go to & Strict \% [95\% CI] & Contain\\
\midrule
Transcript only & --- & 41 [32--51] & 58\\
+ records, model reasons to verdict & model & \textcolor{bad}{18} [12--27] & 67\\
\quad same, conservative prompt & model & \textcolor{bad}{11} [6--19] & 62\\
Extract $\rightarrow$ compute & code & \textbf{85} [77--91] & \textbf{86}\\
\midrule
\multicolumn{4}{@{}l}{\emph{Compute-step control: one extraction, same four fields, catalog in context; only who computes Budget/Timeline differs}}\\
\quad GPT-4o-mini, model computes & model & 43 [34--53] & 60\\
\quad GPT-4o-mini, code computes & code & \textbf{85} [77--91] & \textbf{85}\\
\quad o3-mini, model computes & model & 62 [52--71] & 71\\
\quad o3-mini, code computes & code & \textbf{80} [71--87] & \textbf{81}\\
\quad GPT-5, model computes & model & 64 [54--73] & 78\\
\quad GPT-5, code computes & code & \textbf{81} [72--88] & \textbf{81}\\
\quad Claude Sonnet~5, model computes$^\dagger$ & model & 76 [67--83] & 81\\
\quad Claude Sonnet~5, code computes$^\dagger$ & code & \textbf{80} [71--87] & \textbf{82}\\
\quad Kimi~K2.6, model computes$^\dagger$ & model & 75 [66--82] & 79\\
\quad Kimi~K2.6, code computes$^\dagger$ & code & \textbf{80} [71--87] & \textbf{80}\\
\bottomrule
\end{tabular}
\end{table*}

\begin{figure}[t]
\centering
\resizebox{\columnwidth}{!}{%
\begin{tikzpicture}
\def\barTop{7.55}
\gridaxis
\draw[dashed,soft,line width=.6pt] (41*\axisW/100,-0.2) -- (41*\axisW/100,\barTop);
\draw[dashed,good,line width=.6pt] (85*\axisW/100,-0.2) -- (85*\axisW/100,\barTop);
\barrow{7.10}{bad!70}{43}{\scriptsize 43}
\node[anchor=east,font=\scriptsize] at (-0.15,7.27) {GPT-4o-mini (model)};
\barrow{6.65}{good!80}{85}{\scriptsize\bfseries 85}
\node[anchor=east,font=\scriptsize] at (-0.15,6.82) {GPT-4o-mini (code)};
\barrow{6.05}{bad!70}{62}{\scriptsize 62}
\node[anchor=east,font=\scriptsize] at (-0.15,6.22) {o3-mini (model)};
\barrow{5.60}{good!80}{80}{\scriptsize\bfseries 80}
\node[anchor=east,font=\scriptsize] at (-0.15,5.77) {o3-mini (code)};
\barrow{5.00}{bad!70}{64}{\scriptsize 64}
\node[anchor=east,font=\scriptsize] at (-0.15,5.17) {GPT-5 (model)};
\barrow{4.55}{good!80}{81}{\scriptsize\bfseries 81}
\node[anchor=east,font=\scriptsize] at (-0.15,4.72) {GPT-5 (code)};
\barrow{3.95}{bad!70}{87}{\scriptsize 87}
\node[anchor=east,font=\scriptsize] at (-0.15,4.12) {gpt-5.6-sol (model)};
\barrow{3.50}{good!80}{87}{\scriptsize\bfseries 87}
\node[anchor=east,font=\scriptsize] at (-0.15,3.67) {gpt-5.6-sol (code)};
\barrow{2.90}{bad!70}{76}{\scriptsize 76}
\node[anchor=east,font=\scriptsize] at (-0.15,3.07) {Claude Sonnet 5 (model)};
\barrow{2.45}{good!80}{80}{\scriptsize\bfseries 80}
\node[anchor=east,font=\scriptsize] at (-0.15,2.62) {Claude Sonnet 5 (code)};
\barrow{1.85}{bad!70}{87}{\scriptsize 87}
\node[anchor=east,font=\scriptsize] at (-0.15,2.02) {Claude Fable 5.1 (model)};
\barrow{1.40}{good!80}{88}{\scriptsize\bfseries 88}
\node[anchor=east,font=\scriptsize] at (-0.15,1.57) {Claude Fable 5.1 (code)};
\barrow{0.80}{bad!70}{75}{\scriptsize 75}
\node[anchor=east,font=\scriptsize] at (-0.15,0.97) {Kimi K2.6 (model)};
\barrow{0.35}{good!80}{80}{\scriptsize\bfseries 80}
\node[anchor=east,font=\scriptsize] at (-0.15,0.52) {Kimi K2.6 (code)};
\barrow{-0.25}{bad!70}{80}{\scriptsize 80}
\node[anchor=east,font=\scriptsize] at (-0.15,-0.08) {Qwen 3.8 Max (model)};
\barrow{-0.70}{good!80}{80}{\scriptsize\bfseries 80}
\node[anchor=east,font=\scriptsize] at (-0.15,-0.53) {Qwen 3.8 Max (code)};
\end{tikzpicture}}
\caption{Compute-step control across eight models, four providers.
Each pair holds the extraction fixed and varies only who computes Budget and
Timeline: red bar = model, green bar = code. The grey dashed line at 41 is what the transcript alone gets you; the green dashed line at 85 is where GPT-4o-mini's code arm lands, both from Table~\ref{tab:sameinfo}. Across the eight displayed runs, code did not lower either metric; the observed differences range from $+42$ on GPT-4o-mini to $0$--$+5$ on models that already compute correctly. Two seeds per model on o3-mini and GPT-5 are insufficient to characterize stability, so we report an observed pattern, not a guaranteed performance floor.}
\label{fig:computeControl}
\end{figure}
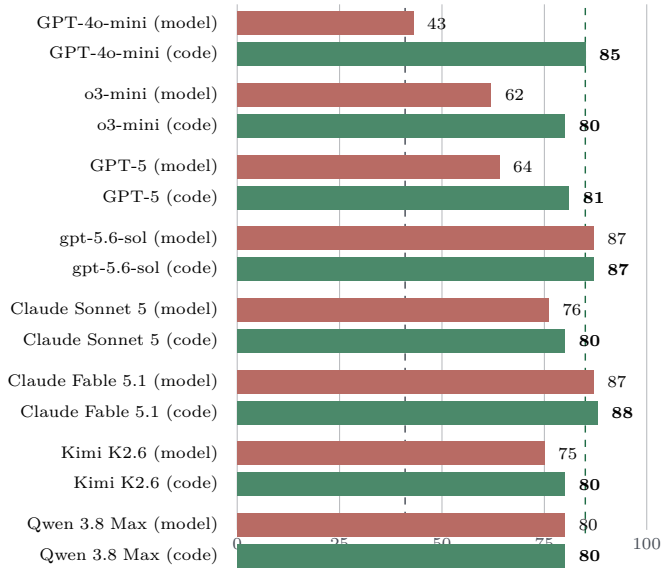

The shape of the ladder is the argument. Chain-of-thought \cite{cot} helps
($41\rightarrow56$) but plateaus, because no amount of reasoning over the
transcript can supply a price that is not present in it. The revealing rung is
the third: providing the model with the missing records and asking it to reason
to an answer reduces strict accuracy to 28\% (GPT-4o, training split). This is
not an information problem, and it is not the credulity of Section~I either; it
is the opposite failure. Handed the price list and policy, the model swings from
believing every claim to distrusting every case: it over-reports Budget and
Timeline and corrupts the Authority and \emph{None} cases it previously answered
correctly. The two metrics make the mechanism explicit. On the same-information
runs of Table~\ref{tab:sameinfo}---one model, all 100 tasks, so nothing is mixed
across readers---giving the model the records \emph{raises} contain
($58\rightarrow67$) while \emph{lowering} strict ($41\rightarrow18$). Recall of
true failures improves, precision collapses, and exact-set grading punishes the
spurious extra factors. The records help the model find failures; they do not
help it stop at the right ones. The effect replicates and is not an artifact of
the inexpensive reader: a repeat run on the GPT-4o-mini gives $41\rightarrow20$ strict and $59\rightarrow71$ contain, and on the GPT-4o the collapse is larger, $53\rightarrow20$ strict against $56\rightarrow68$ contain (95\% intervals
43--62 and 13--29), with Authority falling from 25/27 to 8/27 and \emph{None}
from 15/20 to 6/20. Differences of one to two points between repeated
temperature-zero runs reflect provider-side nondeterminism \cite{ouyang2023}.

The correction is architectural, and we state precisely what it changes. The
model's task becomes extraction into four narrow per-factor fields;
deterministic code then computes exactly two of them---Budget, by summing
quantity$\times$price against the stated budget, and Timeline, by mapping total
unit volume to an installation tier against the required window. Authority and
Need remain the model's own booleans, relayed unchanged; code never decides
them. Two consequences follow. First, one might attribute the gain to
decomposition---asking four narrow questions instead of one holistic
verdict---rather than to computation. We tested this directly by holding
everything but the compute step fixed: one extraction call, the same four
fields, the catalog in context, and from the identical extracted JSON we graded
the model's own \texttt{budget\_fail}/\texttt{timeline\_fail} against code's
recomputation from the same products (Table~\ref{tab:sameinfo}, lower block).
With the model computing, strict accuracy is 43\%; with code computing from the
very same extraction, 85\%. Decomposition alone moves Authority only from 13/27
to 15/27; it reaches 27/27 only once code computes Budget and Timeline
correctly, because the model's spurious Budget and Timeline flags were
contaminating otherwise-correct Authority and \emph{None} sets. The gain is the
computation, not the question format.

Stronger models narrow this gap but do not close it, and the margin
replicates: the o3-mini gives 62\%/80\% (model/code); the frontier
reader 64\%/81\%. Because these models are not deterministic at temperature
zero, we repeated both: reasoning gives 71\%/82\% on a second run, frontier
58\%/79\%, so the model-computes arm swings by 6--9 points between runs while
the code arm moves by 2, and the pooled margins are $+14$ and $+19$. These
models now perform the arithmetic themselves---both match code on Timeline
(9/11 in every run; the GPT-5 scored 11/11 once), and reasoning comes
within two of code on Budget, though frontier still trails there (9--11/19
against 16/19)---so computation is a much smaller part of their loss, and
code's advantage is not uniform across factors. What
remains is over-flagging of qualified deals (\emph{None} 10--19/20 and 14--15/20
across runs, against 20/20 in every code run) and the Authority sets it
contaminates (19--23/27 against 27/27), which code eliminates.

On the second provider the margin nearly closes: Claude~Sonnet~5
computes Budget and Timeline almost as well as code does (16/19 and 7/10
against 17/19 and 7/10) and over-flags almost nothing (\emph{None} 17/18), so
its pair is 76\% against 80\%, and $+2$ to $+4$ over three runs---well inside
the interval. Kimi~K2.6 behaves like Claude, not like GPT-5:
75\% against 80\%, with code's gain again coming from Authority (25/27 to
27/27) and \emph{None} (15/17 to 17/17) rather than from arithmetic. Figure~\ref{fig:computeControl} presents this pattern. We therefore state the claim at its true strength: code's advantage is largest on the GPT-4o-mini ($+42$),
persists at $+14$ to $+19$ pooled over two runs on the frontier and reasoning
models, where it is a precision advantage rather than an arithmetic one, and
narrows to $+2$ to $+5$ on the two models in our set that already compute
correctly and rarely over-flag. What code guarantees on every model is a soundness property, not a lift: it never lowered
either metric. The intervals in Table~\ref{tab:sameinfo} calibrate this: on the efficient
reader the model-versus-code intervals are well separated (34--53 against
77--91), whereas on o3-mini and GPT-5 they meet or barely
overlap at their boundary (52--71 against 71--87; 54--73 against 72--88). Each of those margins is therefore only marginally resolved at $n{=}100$ on its own; what makes it credible is that two independent models reproduce the same sign and magnitude. A paired McNemar test on matched per-case outcomes would settle the arm-to-arm difference at the same $n$; we report the marginal intervals here and release the per-case outcomes for that analysis.

The second consequence is that the code path is \emph{fail-safe against fabrication}: when an extracted product name does not match the catalog, code refuses to flag Budget rather than inventing a verdict from an unknown price. This is a soundness property, not a recall property; a genuine budget failure can still be missed when extraction itself fails.

\subsection{Two Implementation Details}
Two details each affected accuracy more than any change of model. First, tier
direction: the installation policy assigns a timeline tier by \emph{floor}, the
highest volume threshold not exceeded (4 units remain in the 1-day tier; 9 units
fall in the 3-day tier). Encoding it as a ceiling inverted small orders into
false Timeline failures and cost approximately 50 points on the training split.
Second, grader flattening: multi-factor gold answers are stored as a single
comma-joined string, so a naive set comparison scores a correct answer as
incorrect; splitting both sides before comparison recovered several points that
were never model errors.

\begin{table}[t]
\centering
\caption{Per-factor accuracy at the top of the ladder (GPT-4o). Budget and
Timeline are computed by code from the extracted quantities; Authority, Need,
and \emph{None} are the model's own narrow booleans, relayed by code. The
computed factors become answerable; the residual Need misses are model
judgments.}
\label{tab:perfactor}
\small
\begin{tabular}{@{}lcl@{}}
\toprule
Factor & Correct & Answerable from\dots\\
\midrule
Authority & 27/27 & transcript content\\
None (qualifies) & 20/20 & transcript content\\
Budget & 18/19 & catalog prices + computation\\
Need & 12/15 & transcript (subjective)\\
Timeline & 9/11 & install policy + computation\\
\bottomrule
\end{tabular}
\end{table}

Prior to any evaluation, we classified all 22 B2B task types by a rule committed
to a timestamped file before any score was observed: is the answer derivable
from the natural-language content of activity records, without arithmetic over
rows, a numbered rulebook, or data outside the ingested scope? Only two types
qualify; lead qualification is the one we carry to completion because it lies on
the boundary. One caveat applies to the other content type,
\texttt{knowledge\_qa}: CRMArena grades it by token-overlap F1 in the manner of
extractive question answering \cite{squad}, which on its
approximately 50-character reference answers penalizes a correct answer that is
phrased differently. Even a strong model that selects the correct source article
and answers correctly averages about 0.04 F1, so we do not interpret that metric
as an answerability rate.

\section{Where the Approach Ceases to Apply}
A mechanism is only as useful as the boundary of its applicability. To locate
that boundary we pre-specified, in the manner of a pre-registration \cite{nosek2018}, a generalization test: before writing any code
or observing any score, we recorded a prediction that a second task type would
also succeed. The task, \texttt{invalid\_config}, audits a quote against company
regulations and returns the identifier of the violated policy article. Its facts
are already structured and its policy is a short rulebook, so we predicted at
least 85\% accuracy.

The prediction was wrong, for a reason worth stating precisely: the correct
answer cannot be recovered from the information the task provides. The same
violation is marked differently in different tasks. A quote with CloudLink
Designer at 20 units, against a stated limit of 15, is marked a quantity-limit
violation in one task, compliant in another, and a missing-bundle violation in a
third---yet every field that could indicate which rule to apply is identical
across all three. No rule we constructed from the visible fields exceeded approximately 68\%, and three quotes with identical extracted inputs (CloudLink Designer at 20 units against a stated limit of 15) receive three different labels across task instances---compliant, quantity-limit violation, and missing-bundle violation. We do not claim a formal impossibility bound, and we did not exhaustively search rule space; we claim only that no rule we found recovers the labels, and that this failure appears to be a property of the inputs rather than of the approach.

This is not a claim that the benchmark is mislabeled in the sense of
\cite{northcutt2021}. The likely cause is benign:
each quote was constructed to exercise one intended rule, and the answer key
records that intention, which the inputs do not expose. We report the conclusion
provisionally, since we could find no rule that recovers the labels, which is
weaker than proving that none exists. A second candidate,
\texttt{policy\_violation\_identification} (does a case resolution conflict with
a knowledge article?), has the same character: the label is determined neither
by the article's recommended-solution text nor by the one quantitative rule the
articles state, the eligibility window in days between the order date and the
case, whose values for violated and compliant cases of the same issue overlap
completely. We therefore did not build a solver for it, and we count it as a
second underdetermined type rather than as a generalization result in either
direction.

\begin{table}[t]
\centering
\caption{The precondition, stated as a rule.}
\label{tab:precond}
\small
\begin{tabular}{@{}p{0.92\columnwidth}@{}}
\toprule
Extract-then-compute requires a policy that is exactly specified \emph{and}
identifiable from the inputs, not merely documented in prose. Lead qualification
satisfies this; \texttt{invalid\_config} does not, which bounds what any
deterministic method can achieve on it.\\
\bottomrule
\end{tabular}
\end{table}

\section{Related Work}
\textbf{Sycophancy and knowledge conflicts.} Sycophancy \cite{sycophancy, mwe, weisyc} concerns deference to the \emph{user's} stated view. Knowledge conflict \cite{xie2024, pan2023} concerns deference to in-context evidence that contradicts a model's parametric prior. Our setting is neither: the persuasive party is a witness recorded in the context, not the questioner, and the model has no parametric prior about a specific deal's price. Kadavath \emph{et al.} \cite{kadavath2022} argue that models can be trained to know when they do not know; we measure a case where the model has no calibrated signal that the source itself is unreliable, and where the correction is at the pipeline level rather than the model level.

\textbf{Motivated testimony.} The closer neighbor is work on what happens when the text a model is reading was written by someone with a stake in the answer. Knowledge-conflict work \cite{xie2024} shows the model shifts once the source is flagged as unreliable; misinformation-pollution work \cite{pan2023} shows it absorbs the claim anyway when the source has an adversarial motive. Our case is quieter: the source is a sales representative whose incentive to sound optimistic is a stable fact of the role, but the context never flags it that way. We did not try prompts that make the incentive obvious, and that is the natural comparator we owe.

\textbf{Offloading computation to code.} The remedy is not new. Transformers
are unreliable at exactly the multi-step arithmetic the verdict requires
\cite{dziri2023,cobbe2021}. PAL \cite{pal}
has the model write a program as its reasoning trace and a Python runtime execute
it; Chain of Code \cite{coc} extends this to semantic sub-steps an interpreter
cannot run; Toolformer \cite{toolformer} teaches models to call calculators and
other tools; ReAct \cite{react} interleaves reasoning with tool calls.
Extract-then-compute is the same division of labor, specialised: the model
produces structured facts rather than a program, and a fixed, audited
policy---not model-written code---performs the computation. We claim no novelty
for the architecture. Our contribution is the diagnosis of \emph{why} it is
required here (persuasion, not missing information), the controls showing that
supplying the records to the model does not substitute for it, and the isolation
of the computation step as the operative component.

\textbf{Enterprise agent benchmarks.} CRMArena-Pro \cite{crmarena} is one of
several recent benchmarks that place agents inside business software:
$\tau$-bench \cite{taubench} tests policy-following in tool--agent--user
dialogues, WorkArena \cite{workarena} tests web agents on ServiceNow, and
AppWorld \cite{appworld} tests interactive coding agents across applications.
These report aggregate task success; we instead take one task type, characterize its failure mechanism, and use pre-specified controls to isolate the causal component. Retrieval-augmented generation \cite{rag,dpr} is the standard means of bringing records into context; agent memory systems \cite{genagents,memgpt,mem0} extract and consolidate them. Neither line prescribes whether the ultimate verdict is drawn from the model's reading of the retrieved text or from a deterministic computation over its extracted contents. Our controls quantify the accuracy consequence of that choice.

\section{Limitations and Conclusion}
The central finding rests on \emph{one task type} (lead qualification, $n{=}100$) from \emph{one benchmark} (CRMArena-Pro), and the single pre-specified generalization test returned a negative result. Accuracy is reported against one grader, and the cost figures pertain to a
101-account corpus and a single embedding model. The scope is narrow by
construction: B2B, single-turn, and synthetic. We do not evaluate multi-turn
interaction, business-to-consumer settings, or the benchmark's privacy-rejection
tasks, and synthetic data flatters a deterministic approach that is more brittle
than an agent to the noisy, custom-field reality of a production system. One
caution is specific to the finding: although the effect holds across seven
models from four providers, spanning current and prior-generation OpenAI, Anthropic (Sonnet~5, Fable~5.1), Moonshot, and Alibaba, with
open-weights families (Table~\ref{tab:sweep}), the contradiction set is small
($n=31$), so the table establishes a direction rather than precise rates;
confirming them on a larger set remains future work. The compute-step control is reported on five models but rests on $n=100$ per arm, and its margin on the strongest models is resolved only at the boundary of the intervals. On Claude Sonnet 5 and Kimi K2.6 the code arm is within the model arm's confidence interval; the honest reading is that on models that already compute correctly and rarely over-flag, code shows a consistent direction rather than a measured lift. We report the pattern as a directional soundness result and do not claim uniform superiority.

Three conditions must hold before any unsupervised deployment, and none is
established here. First, access control: a compiled solver that issues SQL
directly bypasses field-level security and record sharing, and in a production
system the same query must execute under the caller's permissions. Second,
abstention: we report point accuracy rather than a calibrated ``escalate when
uncertain'' signal \cite{kadavath2022}, and acting on an 84--90\% answer without such a gate will
misroute the tail. Third, trust in extraction: the extractor remains a language
model, and an incorrectly extracted quantity---a hallucination in the strict
sense \cite{ji2023}---yields a confidently incorrect computed answer, so the
structured facts require validation. The mechanism moves the error from reasoning into extraction; it does not remove it. In particular, the extractor itself is a language model reading the same optimistic transcript, and a systematic bias in extracted budgets or timelines would not be caught by any downstream code. Our per-factor decomposition (Table~\ref{tab:perfactor}) shows Budget and Timeline extraction agree with code's re-derivation from the same JSON, but this is not a proof of extractor calibration under adversarial conditions.

We contribute a diagnostic method. When a model answers a question over records that include an in-context assertion by an incentive-misaligned witness, the danger is not that it cannot reason but that it treats the assertion as evidence, and the failure has a specific signature (recall up, precision down) that supplying the records to the model does not correct. Our bucket analysis separates persuasion from missing information; our same-information control shows the direction of the effect; our compute-step control isolates the operative component. The pattern is a consistent direction rather than a uniform lift: on models that already compute correctly and rarely over-flag, the gain narrows to within confidence intervals. The precondition for the correction is that the policy be exact and identifiable from the inputs. When it is not---as our pre-specified generalization test shows for \texttt{invalid\_config}---no method, in-context or in-code, exceeds the ceiling the inputs impose. The question to ask before reaching for a larger model is which of these two regimes the task is in. If the policy is written down precisely and the inputs make it identifiable, take the arithmetic out of the model and put your effort into the extraction step. If the policy is not written down precisely, no model and no amount of in-context grounding will help; either write the policy or accept what the inputs can support.

\appendices
\section{A Traced Response, Both Paths}
The same lead-qualification request is executed two ways. The values are
illustrative.

{\scriptsize
\textit{A.1\quad Extract-then-compute}
\begin{verbatim}
[request]  qualify this lead? if not, which BANT factors fail?
           wants 9x "PulseSim Pro", budget ~$4,000,
           needs it within 1 day, "I'll run it by my VP";
           rep: "that fits your budget, we install in a day"

[extract]  inexpensive LLM ->
           {"products":[{"name":"PulseSim Pro","qty":9}],
            "stated_budget":4000,"required_days":1,
            "authority_fail":true,"need_fail":false}

[lookup]   catalog["PulseSim Pro"] = $599.99/unit
           policy: 9 units -> 3-day tier (5+ -> 3 days)

[compute]  9 x 599.99 = $5,399.91 > $4,000  -> FAIL
           3-day tier > 1 day required       -> FAIL
           authority defers to VP            -> FAIL
           need genuine                      -> ok

[answer]   Authority, Budget, Timeline   CORRECT
           1 inexpensive call, ~$0.0002/query
\end{verbatim}

\textit{A.2\quad Tool-using agent---reason in the loop}
\begin{verbatim}
[step 1]  plan: "inspect the schema first"       . model
[step 2]  describe Lead, Product2, Pricebook      . model+tool
[step 3]  write SOQL for products and prices      . model
[step 4]  execute query, read rows                . tool
[step 5]  budget needs a pricebook join -> re-run . model+tool
[step 6]  reason over the numbers -> answer        . model

[answer]  "None - the lead is qualified"   INCORRECT
          (accepted "fits your budget" / "install in a day")
          6 GPT-5 calls, ~$0.08/query (est.)
\end{verbatim}
}

\section*{Reproducibility}
All accuracy figures are measured. The evaluation harnesses, the pre-specified task classification, the generalization-test pre-specification (written down before any code was run or score observed, though not deposited with an external timestamping service), the extraction prompts, and the per-run result files are included as ancillary files with this submission. Model identifiers used: OpenAI \texttt{gpt-4o-mini}, \texttt{gpt-4o}, \texttt{o3-mini}, \texttt{gpt-5}, \texttt{gpt-5.6-sol}; Anthropic \texttt{claude-sonnet-5}, \texttt{claude-fable-5-1}; Moonshot \texttt{kimi-k2.6}; Alibaba \texttt{qwen/qwen3.8-max-0902} (via OpenRouter). Kimi and Qwen permit only temperature~1; all other extractions run at temperature~0. The data split is the CRMArena 80/20 split (seed 42). This is an independent analysis using the CRMArena-Pro B2B tasks and data; it is not affiliated with or endorsed by the benchmark's authors.

\textbf{Data provenance and license.} Interactive access to the benchmark's live Salesforce organization is restricted, so the B2B records (transcripts, catalog, policies) were read from the benchmark's offline SQLite materialization (\texttt{crmarenapro\_b2b\_data.db}) as redistributed in public GitHub mirrors of the CRMArena code; the task set is the Hugging Face release. CRMArena-Pro is licensed CC~BY-NC~4.0. This is non-commercial research, and no benchmark records are redistributed with this paper.

\textbf{Conflict of interest.} The author is employed by AmpUp, which sells sales software that uses the extract-then-compute architecture this paper argues for.

\end{document}